\documentclass[letterpaper, 10 pt, conference]{ieeeconf}  

\IEEEoverridecommandlockouts                              

\usepackage{amsmath} 
\usepackage{amssymb}  
\usepackage{graphicx}
\usepackage{algorithm}
\usepackage{algpseudocode}
\usepackage{bm}
\usepackage{courier}
\usepackage[nolist]{acronym}
\usepackage{tikz}
\usepackage{textcomp}
\usepackage{hyperref}
\usepackage{lipsum}

\algnewcommand\algorithmicforeach{\textbf{for each:}}
\algnewcommand\ForEach{\item[ \algorithmicforeach]}

\title{\LARGE \bf
Event-Based high-speed low-latency fiducial marker tracking
}

\author{Adam Loch $^{12*}$, Germain Haessig$^{1*}$, Markus Vincze $^{2}$
\thanks{$^{*}$These two authors contributed equally}
\thanks{$^{1}$Authors are with the AIT Austrian Institute of Technology, Vienna, Austria \texttt{\{{adam.loch,germain.haessig}\}@ait.ac.at}}%
\thanks{$^{2}$Authors are with the Technical University of Vienna, Automation and Control Institute (ACIN), Vienna, Austria \texttt{vincze@acin.tuwien.ac.at}%
}%
\thanks{We are grateful to Bernhard Blaschitz and Philipp Schneider (AIT) for insightful research discussions, Thomas Weingartshofer and Christian Hartl-Nesic (ACIN) for collaboration in the process of ground truth acquisition.
}%
}

\newcommand\copyrighttext{%
  \footnotesize This work has been submitted to the IEEE for possible publication. Copyright may be transferred without notice, after which this version may no longer be accessible.}
\newcommand\copyrightnotice{%
\begin{tikzpicture}[remember picture,overlay]
\node[anchor=south,yshift=10pt] at (current page.south) {\fbox{\parbox{\dimexpr\textwidth-\fboxsep-\fboxrule\relax}{\copyrighttext}}};
\end{tikzpicture}%
}

\begin{document}

\maketitle
\thispagestyle{empty}
\pagestyle{empty}

\begin {acronym}
\acro {6dof} [6-DOF] {6 degrees of freedom}
\acro {fps} [FPS] {Frames per second}
\acro {pnp} [PnP] {Perspective-n-Point}
\end {acronym}

\copyrightnotice
\begin{abstract}

Motion and dynamic environments, especially under challenging lighting conditions, are still an open issue for robust robotic applications. In this paper, we propose an end-to-end pipeline for real-time, low latency, 6 degrees-of-freedom pose estimation of fiducial markers. Instead of achieving a pose estimation through a conventional frame-based approach, we employ the high-speed abilities of event-based sensors to directly refine the spatial transformation, using consecutive events. Furthermore, we introduce a novel two-way verification process for detecting tracking errors by backtracking the estimated pose, allowing us to evaluate the quality of our tracking. This approach allows us to achieve pose estimation at a rate up to 156~kHz, while only relying on CPU resources. The average end-to-end latency of our method is 3~ms. Experimental results demonstrate outstanding potential for robotic tasks, such as visual servoing in fast action-perception loops.

\end{abstract}

\section{INTRODUCTION}

Recent advancements in event-based sensor technology open new exciting possibilities for applications of event-based vision in robotics. Manufacturers (e.g. Prophesee~\cite{prop_new}, Samsung~\cite{sam_new}, CelePixel~\cite{cele}) drastically advanced sensor quality in the past two years. Among the most important improvements lie an increased resolution, as well as a decreased noise level. The current generation of those cameras provides resolutions much closer to conventional frame-based sensors (up to 720p). Given the main advantages of event-based sensors, namely high dynamic range and high temporal resolution, they can be useful for many high-speed robotics applications, thus breaching the gap in real-time fast action-perception loops.
\par
Object pose estimation is essential in a variety of robotic tasks. The performance of those algorithms has a significant impact on the overall outcome of robotic solutions~\cite{benchmarking_pose_estimation}. Current state-of-the-art algorithms~\cite{bop} improved substantially over the last few years. However, most of them are not well suited for dynamic scenes due to high execution time and motion blur in the signal. Additional delay introduced by the acquisition and recognition steps makes them unsuited for real-time scenarios.
\par

Nonetheless, not every task requires the pose estimation of an object. In some cases, only the reference frame is required; the problem can then be simplified by employing visual markers, in order to get fast and reliable pose estimations. Newer markers, such as AprilTags~\cite{apriltag}, STags~\cite{stag}, CCTags~\cite{cctag}, and currently the most popular ArUco markers~\cite{aruco}, show improvements in various fronts: speed, robustness, and precision. To obtain a reference position in dynamic scenes, however, motion blur, challenging lighting conditions and estimation speed have to be taken into account. 

\par
\begin{figure}[]
\centering
\includegraphics[scale=0.3]{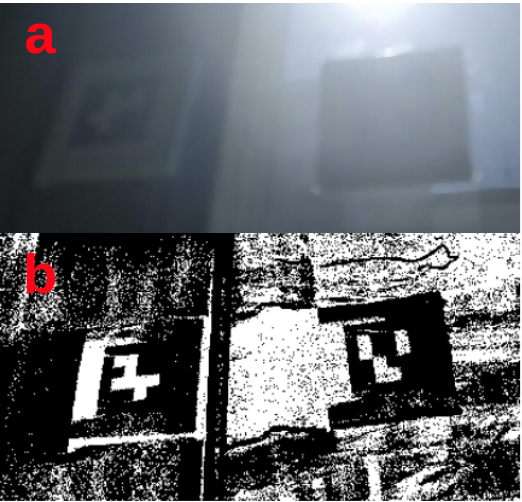}
\caption{Comparison of (a) an RGB camera frame and (b) event-based sensor events accumulation, under challenging lighting conditions. The events are represented by the frame, which indicates the polarity of the last event at every pixel. Due to the challenging lighting conditions and global exposure of standard RGB cameras, one can barely distinguish the markers in the RGB frame. In opposition, with every pixel of the event-based sensor being independent, such conditions do not affect an event-based approach.}
\label{sun}
\end{figure}

In this paper, we propose a real-time method for ArUco marker tracking and refinement using an event-based sensor. In the presented method, tracking is achieved through direct updates of the \ac{6dof} pose using consecutive events instead of an accumulated representation~\cite{markerdetectioneb}. To control and verify the quality of the tracking, we introduce a two-way verification process by backtracking the estimated pose. To generate the initial guess, we rely on the well-known method for ArUco marker detection~\cite{aruco}, using a binned representation of the events. Given the high dynamic range and temporal resolution of the sensor, our system can detect and estimate poses of markers in challenging lighting conditions (Figure~\ref{sun}) as well as during fast motion. 

The main contributions of this paper can be summarized in the following points:
\begin{itemize}
    \item An efficient method of pose tracking for fiducial markers using an event-based sensor.
    \item A method to estimate the quality of tracking using two-way verification.
\end{itemize}

To strengthen our position, we include a comparison with an OptiTrack~\cite{optitrack} ground truth, in order to assess accuracy in well-established robotic solutions.

\par

In the following, Section~\ref{relwork} describes the current state-of-the-art in the field of pose estimation, fiducial marker detection using RGB cameras and event-based sensors. Details of the algorithm and implementation are given in Section~\ref{methodology}. Section~\ref{experiments} describes the experimental methodology and presents results. Finally, Section~\ref{conclusion} summaries the work and shows possible extensions and applications.
\par

\section{RELATED WORK}
\label{relwork}
\subsection{Fiducial markers}
Fiducial markers are a well-known solution for pose estimation tasks, current applications varying from indoor localization~\cite{indoor}, augmented reality~\cite{artoolkit}, swarm robotics~\cite{swarms} and many more. 
One of the most popular marker type is the square binary marker. Those markers use edges and corners detectors to extract the outer shape of the object and encode their unique identification using high contrast (typically black and white) cells. The earliest notable type of markers, proposed in 2005, are ARTags~\cite{artag}. The presented approach uses grayscale images to detect the markers by extraction of the edges and grouping them into a form of quadrilaterals. Proposed improvements also include a verification method using checksum and error correction algorithms.

The next significant improvement was achieved with the introduction of the ArUco marker~\cite{aruco}, which generates the markers pattern in a way to reduce the similarity distance between different IDs.
However, this approach requires high-quality images (without challenging lighting conditions or motion blur) in order to successfully detect and recognize the marker. Those limitations were discussed and partially solved in previous work, employing deep learning methods to improve robustness~\cite{robustaruco},~\cite{robustaruco1}.
Experimental results showed major improvements, under challenging conditions, in terms of detection quality but at the cost of an increased processing time.

\begin{figure*}[t]

\includegraphics[scale=0.5]{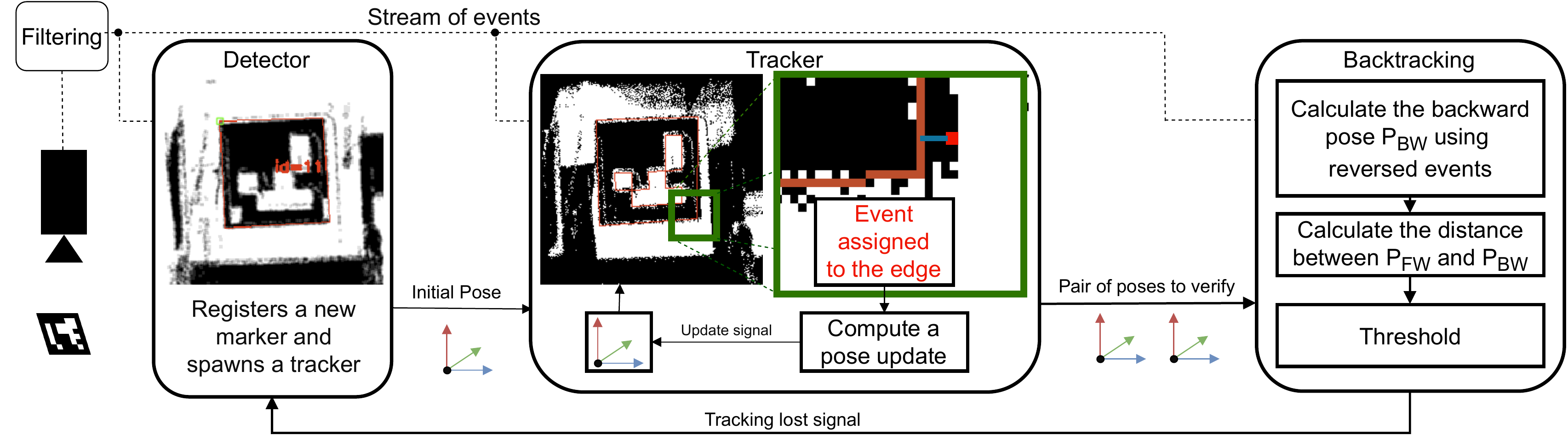}
\caption{Overview of the general pipeline. The detector initializes a new tracker for every newly detected marker. When the distance between the tracked and backtracked poses is greater than a given threshold, the detector deregisters the given tracker.}
\label{gen}
\end{figure*}
\par

While square markers depend on corner detections, which may be hard under dynamic conditions, circulars markers can rely on the estimation of the center of mass. A solution to that problem was proposed by~\cite{blurresistant}, reducing the influence of the motion blur on the detection performance using the recognized blur direction. However, it was not possible to estimate the \ac{6dof} based on a single tag. Some of the proposed circular markers (arrays of circles), RuneTag~\cite{runetag} and Pi-Tag~\cite{pitag} showed improved performance in noisy, blurred scenes. However, the performance of such markers, due to the small size of a single circle, drops significantly while the distance increases.
\par
To the best of our knowledge, the most recent state-of-the-art markers are the TopoTags~\cite{topotag}. This new tag version is more robust to noise and improves the precision of the detected poses, at the cost of a reduced speed (up to 33 \ac{fps}).

\subsection{Event-based vision}
Event-based approaches can be divided into two sub-types: based on accumulated event representation (similar to frames) or event-by-event processing. The first group refers to approaches that accumulate, over time, events in a frame, in order to use it similarly as RGB frames in standard computer vision algorithms. The main advantage of that approach is an improved signal-to-noise ratio. Accumulated representations are used in cases where a global context is required. The frame-based algorithms introduce an additional processing delay due to the accumulation scheme. The recent advancements in this field include end-to-end detection pipeline~\cite{detection1mpix}, iterative pose estimation using RGB data, enhanced with depth (RGB-D) and/or event-based sensor~\cite{rgbde}, as well as image reconstruction~\cite{reconstruction}.

\par
In opposition to the previously mentioned approach is the event-by-event paradigm, focusing on the online processing of the events. It was shown that it can be applied to multiple different tasks, such as filtering methods~\cite{filtering}, algorithms based on particle filter~\cite{particle_filter}, SLAM algorithms~\cite{slam} or pose tracking~\cite{ebpose}. All of these methods rely on an iterative update of a current estimation.
\par
More recently, in~\cite{markerdetectioneb} was introduced a detection method for binary square fiducial markers. The presented results highlighted possible advantages of the event-base approach in cases of high motion blur. However, their detection algorithm highly depends on the direction of the motion and the processing time required for a resolution of 128x128 pixels is worse than the RGB based detection~\cite{aruco}. Moreover, the mentioned paper does not include the pose estimation.
\par

\section{Methodology}
\label{methodology}

Our proposed solution can be divided into three major stages. First, we operate a marker detection; the tracker is initialized using the pose obtained from the corners of the detected marker for every new ID. Then, we assign each event to the closest edge which is assumed to have generated the event. For every assigned event, an incremental update of the \ac{6dof} (translation and rotation) is calculated. To verify the pose, detect a loss of tracking and determine the tracking quality, we employ a reverse tracking approach. A general overview is presented in Figure~\ref{gen}. 
\par
To track markers, we are solely relying on the stream of events. Every event $e_k$ is defined as $e_k = (\bm{u}_k, t_k, p_k)^T$, where $\bm{u}_k = (x_k, y_k)^T $ represents the location of the event on the focal plane, $t_k$ its camera timestamp (time of generation) and $p_k$ its polarity, signaling an increase or decrease of luminosity.

\subsection{Preprocessing}
In order to minimize delays and required computational resources, we limit the number of events by using a noise filter. This spatiotemporal filter reduces the number of events by filtering out the ones which are not supported by another neighborhood event within a given period of time. Here, we considered a time constant of $2000 \mu$s.

\subsection{Detection}
\label{sec:buffer}
\begin{figure}[]
\centering
\includegraphics[scale=0.18]{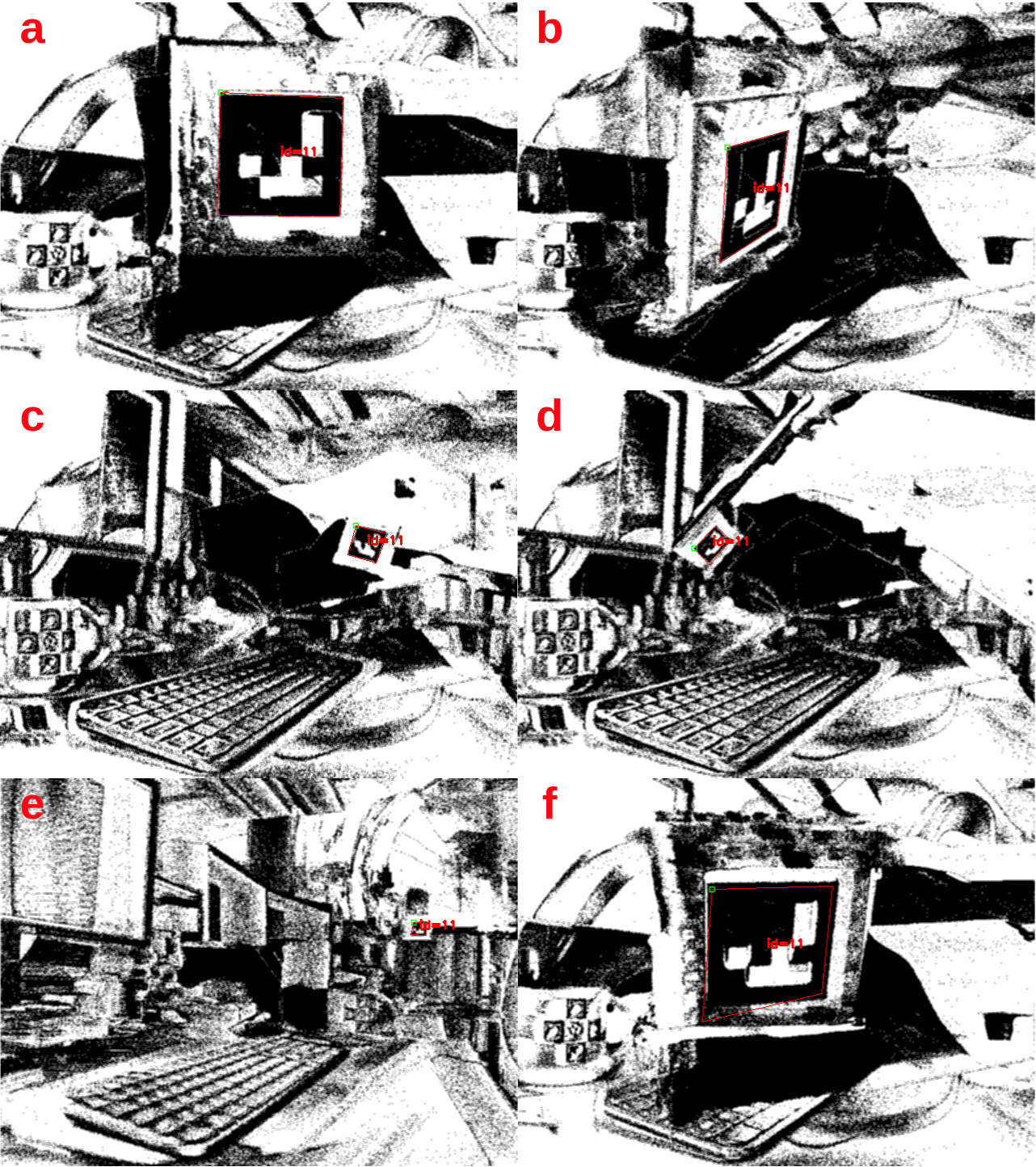}
\caption{Examples of positive detections: a) Close-range without rotation, b) Close-range with high rotation angle, c) Mid-range $\sim1$m,  d) Mid-range with high rotation angle, e) Far-range $\sim2$m f) Example of detections with a misalignment due to noise.}
\label{detection}
\end{figure}

The initial marker detection is achieved following a standard frame-based approach~\cite{aruco}. Exploiting the highly contrasting colors of the ArUco marker allows for a simple event representation. In order to avoid facing issues while determining an accumulation time, our frame aggregates the polarity of the last event for every pixel. Such representation does not require synchronisation, and the current state can be asynchronously updated with every incoming event. Such event frames are presented in Figure~\ref{detection}.
\par
The standard detection algorithm~\cite{aruco} works well with this particular frame representation, even with the background noise generated by our sensor. The detector works asynchronously to the tracking algorithm. Only the detected candidates with an ID which is not currently being tracked are taken into account. In this paper, we are not focusing on the performance of the detection algorithm. In Figure~\ref{detection} some examples of positive detections are presented. All the intermediate events between the start of the detection and the detection output are stored and passed to the initialized tracker in order to avoid event loss.

\par
Once the marker has been detected and identified, its initial pose is estimated via the standard \ac{pnp} solver (based on the Levenberg-Marquardt optimization procedure), using the corners of the detected marker and the camera's known intrinsic matrix $K$. 

\subsection{Tracking}
Our tracking algorithm follows previous work on iterative tracking methods, where, for every event, a transformation update is calculated based on the current estimation and the location of the incoming event. This tracking of a known 3D shape was presented in~\cite{ebpose} and later extended in the~\cite{ebpnp}.

\subsubsection{Initialization}
 Using the known marker 3D model, we can define the vertices' coordinates $\bm{V}_i = (X_i,Y_i,Z_i)^T$. We can then calculate the position $\bm{v}_i$ of every vertex on the camera's focal plane, using the pinhole camera model projection. We define the 3D position of the segment $\bm{S}_{i,j}$ between two 3D vertices $\bm{V}_i$ and $\bm{V}_j$. The marker pose is initialized by the detector output (See~\ref{sec:buffer}). We will be referring to the initial pose as the rotation and translation pair ($\bm{R}^0, \bm{T}^0$)

 \subsubsection{Segment matching}
 For every incoming event $e_k$, we compute its distance to every segment $\bm{S_{i,j}}$. We then select the closest segment with a minimal distance $d_{i,j}(\bm{u}_k)$ defined as follows:
 
  \begin{equation}
     d_{i,j}(\bm{u}_k) = \frac{\lVert (\bm{u}_k - \bm{v}_i) \times (\bm{v}_j - \bm{v}_i) \rVert}{\lVert \bm{v}_j - \bm{v}_i \rVert}
 \label{eq:1}
 \end{equation}
 
 If the distance is higher than a given threshold (tipical 2 pixels), the event is discarded. We can then define the line of sight $\bm{M}_k$ (Equation~\ref{eq:2}) of an event using the known camera matrix $K$. This line of sight can be seen as a 3D line originating from the optical center of the camera and passing through the location of the event $\bm{u}_k$ on the focal plane.
   \begin{equation}
     \bm{M}_k=K^{-1}\binom{\bm{u}_k}{1}
 \label{eq:2}
 \end{equation}
 
 We define the point $\bm{F}_k$ as the 3D point which generates the event $e_k$. We can also define its closest point on the 3D model as a $\bm{E}_k$. Let's assume that the event $e_k$ was induced by the transformation moving $\bm{E}_k$ to $\bm{F}_k$. Both of those values can be computed as~\cite{ebpnp}:
 
\begin{equation}
    \bm{F}_k=\alpha_1\bm{M}_k
 \label{eq:3}
\end{equation}
\begin{equation}
    \bm{E}_k = \bm{V}_j + \alpha_2\bm{S}_{i,j}
 \label{eq:4}
\end{equation}
where $\alpha_1$ and $\alpha_2$ are scalar parameters. To determine $\alpha_2$, we compute the vector $\bm{E}_k - \bm{F}_k$, which is perpendicular to the line of sight $\bm{M}_k$ and to the $\bm{S}_{i,j}$. This results in:
\begin{equation}
    \begin{pmatrix}
-\bm{M}^T_k\bm{M}_k & \bm{M}^T_k\bm{S}_{i,j}\\
-\bm{M}^T_k\bm{S}_{i,j} & \bm{S}^T_{i,j}\bm{S}_{i,j}
\end{pmatrix}
    \begin{pmatrix}
\alpha_1 \\
\alpha_2
\end{pmatrix}
=
    \begin{pmatrix}
-\bm{V}_j^T\bm{M}_k \\
-\bm{V}_j^T\bm{S}_{i,j}
\end{pmatrix}
\label{eq:5}
\end{equation}

As indicated in~\cite{ebpose}, the small transformation assumption from $\bm{E}_k$ to $\bm{F}_k$ holds only due to the high temporal resolution of the signal.

\subsubsection{Translation update}
After matching the generated event to the 3D edge, the last step comprises calculating the iterative update of the pose induced by this event. Due to the high event-rate generated by the sensor (up to 20 Mega Events per second, MEv/s), the pose of the object cannot be updated with every event. To improve performance, we update the pose every $N$ consecutive events. We will denote the current estimation as ($R^*, \bm{T}^*$). 

\begin{figure*}[t]
\centering
\includegraphics[width=\textwidth]{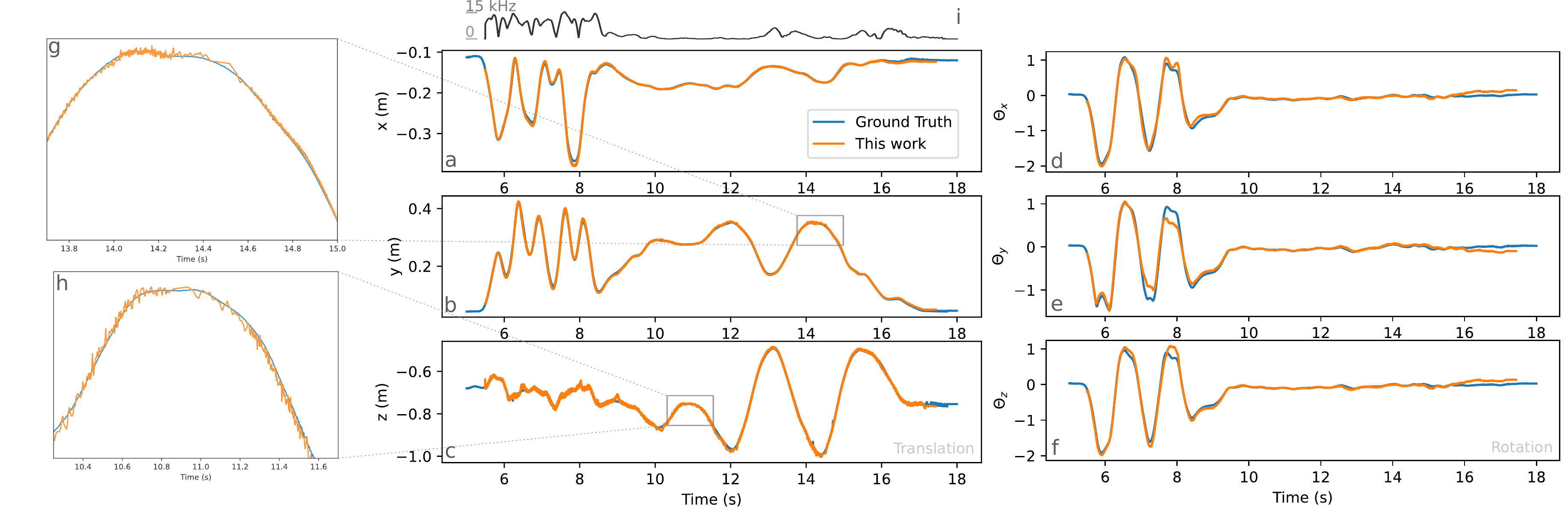}
\caption{\ac{6dof} comparison to ground truth data. Ground truth data was recorded using an external positioning device (OptiTrack). (a-c) Translation values for the 3 axes. The average translation error achieved here is $3.76 \text{mm}$. (d-f) Rotation values for the 3 axes. (g-h) Zoom on 2 specific regions. (i) Instantaneous update rate. Due to the asynchronous nature of the sensor, this update rate is not constant. For this example, we achieve an update rate of up to $15~\text{kHz}$}
\label{optitrack}
\end{figure*}

\par
We can define the line of sight projection matrix $L_k$ as follows:
\begin{equation}
    L_k = \frac{\bm{M}_k\bm{M}_k^T}{\bm{M}_k^T\bm{M}_k}
\label{eq:6}
\end{equation}
\par
Then, we can define the optimal update $\Delta \bm{T}_k$ to the translation component $\bm{T}^*$, for a given rotation $R^*$ as follows~\cite{lu_proofs}:
\begin{equation}
    \Delta \bm{T}_k(R^*,\bm{T}^*)=A_k^{-1}\bm{B}_k
\label{eq:7}
\end{equation}
Where $A_k$ and $\bm{B}_k$ are approximated, such as in~\cite{ebpnp}:
\begin{equation}
    A_k \approx w_0(I_3-L_k)+(1-w_0)A_{k-1}
\label{eq:8}
\end{equation}
\begin{equation}
    \bm{B}_k \approx w_0(L_k-I_3)\bm{F}_k+(1-w_0)\bm{B}_{k-1}
\label{eq:9}
\end{equation}

$A_{k-1}$ and $\bm{B}_{k-1}$ are previous estimates, and $w_0$ is a tune-able hyper-parameter (typical $0.1$). In comparison to the method presented in~\cite{lu_proofs}, our approach can be used in real-time operations.
\par
Every $n$ steps (update rate parameter, typical $100$), the estimation of the translation $T^*$ is updated in the following way:
\begin{equation}
    \bm{T}^*=\bm{T}_{k-1}+\lambda_T\Delta \bm{T}_k
\label{eq:10}
\end{equation}
where $\lambda_T > 0$  is an update rate for the translation component (typical $1.4$).

\subsubsection{Rotation update}
The iterative update of the rotation component proposed in~\cite{ebpnp} is inspired by the virtual mechanical system with the only possible rotation in its origin point. It can be represented as a spherical joint. Using the system of virtual springs to stabilise it, the resulting torque can be approximated as follows:
\begin{equation}
     \Gamma \approx  w_0R^*\bm{F}_i\times(L_k-I_3)\bm{F}_i^*+(1-w_0)\Gamma_{k-1}
\label{eq:11}
\end{equation}

In the same way as for the translation update, the estimated rotation is updated every $n$ steps using (typical $100$):
\begin{equation}
    r=\lambda_r\bm{\Gamma_k}
\label{eq:deltar}
\end{equation}
where $\lambda_r > 0$ is an update rate for the rotation (typical $0.003$). We then can define the rotation matrix $\Delta R_k$ corresponding to the rotation $r$ and update directly the rotation estimation $R^*_k$ as follows:
\begin{equation}
    R^*=\Delta R_kR_{k-1}
\label{eq:12}
\end{equation}

\subsection{Backtracking}
To assess the tracking quality, we introduce a two-way verification process by backtracking the estimated pose. To validate the performance of the tracker between update $n$ and $n+k$, given the initial transformation $P_n = (\bm{T}_n, R_n)$, we introduce a forward ($P_{FW}$) and backward pose ($P_{BW}$). $P_{BW}$ is computed as the transformation originating from $P_{FW}$ and being induced by the events between $n$ and $n+k$, in the reverse order:
\begin{align}
    P_{FW} &= P_n + \Delta P_{FW} \\
    P_{BW} &= P_{FW} + \Delta P_{BW} 
\end{align}

We can then calculate the translation error between these two poses ($P_{FW}$ and $P_{BW}$) using the projections of the center of the marker:

\begin{equation}
    d_T(n,n+k) = |x - x'| + | y - y'|
\label{eq:13}
\end{equation}
where $(x, y)$ (respectively $(x', y')$) are the projected coordinates of the center using the forward (resp. backward) tracking. For the rotational error, we calculate the sum of errors for every axis using Euler angles.
\begin{equation}
    d_R(n,n+k) = |\alpha - \alpha'| + |\beta - \beta'| + |\gamma - \gamma'|
\label{eq:14}
\end{equation}
where $(\alpha, \beta, \gamma)$ (respectively $(\alpha', \beta', \gamma')$) are roll, pitch, heading angles estimated using the forward (resp. backward) tracking.
Then, using the threshold values $\epsilon_T$ (typical $5$) and $\epsilon_R$ (typical $0.15$), we can then check if the tracking was lost. The $k$ value was determined empirically, and its value was set to $100$ updates.

\subsection{Performance optimization}
To reduce processing time and delays, all the events are processed in an online fashion. We relied on the open source framework OpenEB~\cite{openeb} for low-level camera communication. To achieve real-time performance, the processing pipeline was divided into 4 independent threads. We implemented our own inter-thread communication. All the threads are connected using a shared queue as an event buffer. The detector runs asynchronously to the event stream pulling, with the current state of the representation frame. All the active trackers are running in parallel to the detection loop.

\section{Experiments}
\label{experiments}
To characterize and quantify different aspects of the performance of our algorithm, we conducted three different experiments. All of them used the Prophesee Gen3 EVK (resolution 640x480)~\cite{evk}, which in comparison to the sensors used in~\cite{ebpnp},~\cite{ebpose},~\cite{markerdetectioneb}, generates much higher event rate~\cite{survey}. We used a 8mm 1:1.3 lens. The optical system was calibrated using the pinhole model. All the tests were performed using a computer with a high-end CPU (Intel i7-9800X) and 64GB of RAM. 
\subsection{Accuracy}
In order to measure the accuracy of our approach, high-speed ground truth signal was obtained using an OptiTrack system~\cite{optitrack}. This motion capture system allows to capture the \ac{6dof} pose of a rigid body up to 360Hz. In our setup, infrared markers were tracked using 6 OptiTrack Prime 17W cameras. To reduce the noise generated by the blinking infra-red LEDs, we mounted an infra-red filter in front of our optical system. However, due to the filter properties, a slight noise was still noticeable in the signal.
\par
In Figure~\ref{optitrack}, we can see the performance of our tracking under fast motion. As we can observe, the performance of the tracking algorithm degrades with the increased distance from the camera. Due to the limited resolution of the sensor, the operating distance (for our optical configuration) is up to approximately 1,75m. The average translation error, for the shown trajectory, is $3.76 \pm 1.6 \text{mm}$. 

As our approach estimates the \ac{6dof} pose in an asynchronous manner, we performed a linear interpolation of our data in order to temporally align the two data flux.

\subsection{Efficiency}
\begin{figure}[]
\centering
\includegraphics[width=0.5\textwidth]{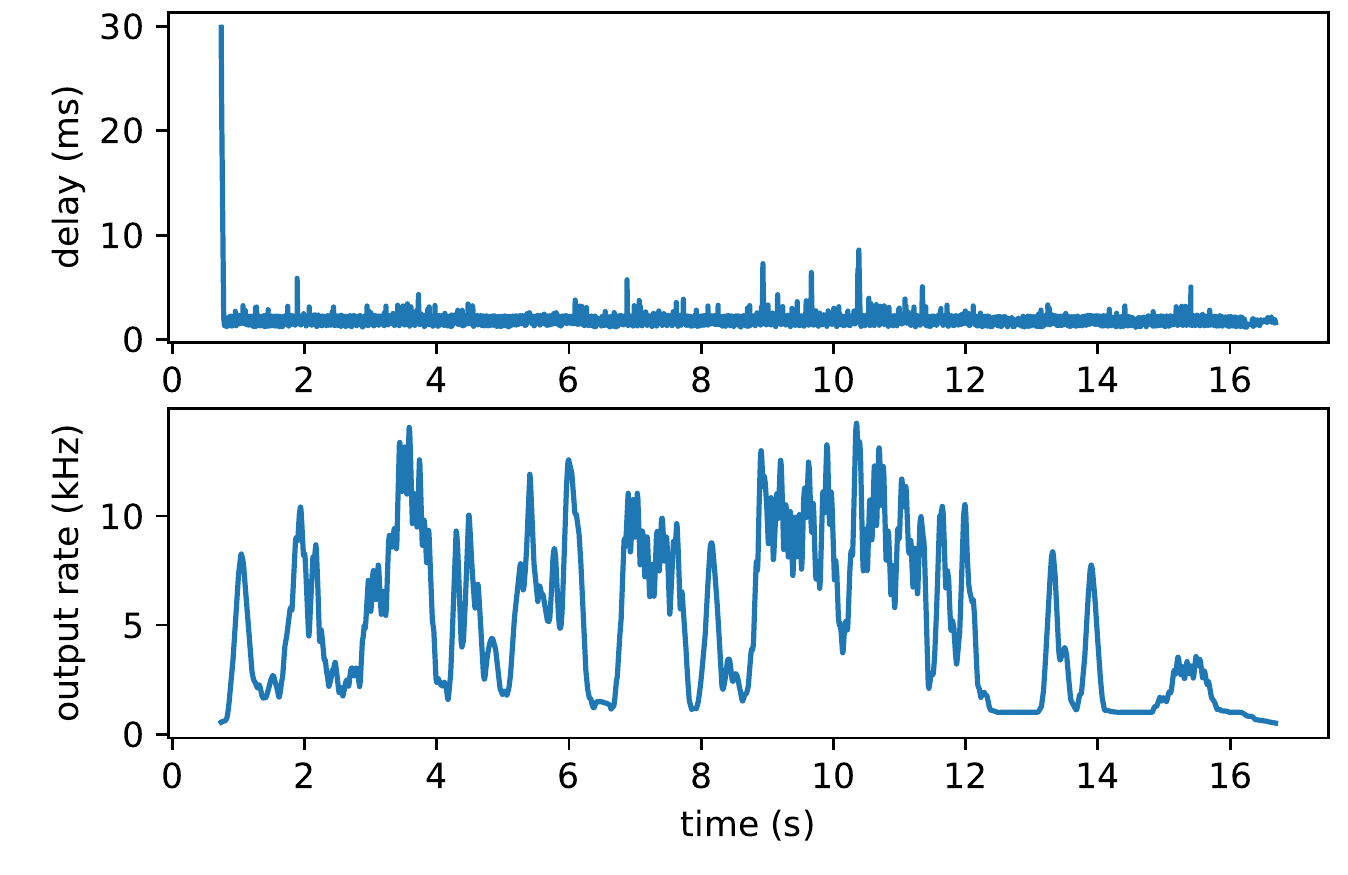}
\caption{End-to-end delay and output rate. The initial delay is the result of the initialization time required and buffered events during the detection loop. The average delay is equal to $2.058 \pm 0.551$ ms.}
\label{delay}
\end{figure}
To measure the end-to-end delay of the whole pipeline, we used an external trigger system in order to synchronise the internal clock of the camera and our computer. Measured processing time includes camera processing delay, transport and processing time. The measured delay and the output rate are presented in Figure~\ref{delay}.
\par
The initial delay of the tracker is due to the initialization time, as well as the processing time required to process the buffered events such as explained in Section~\ref{sec:buffer}. Average delay is here $2.058 \pm 0.551$ ms. 

\subsection{Tracking quality estimation and fail cases}
\begin{figure}[]
\centering
\includegraphics[width=0.5\textwidth]{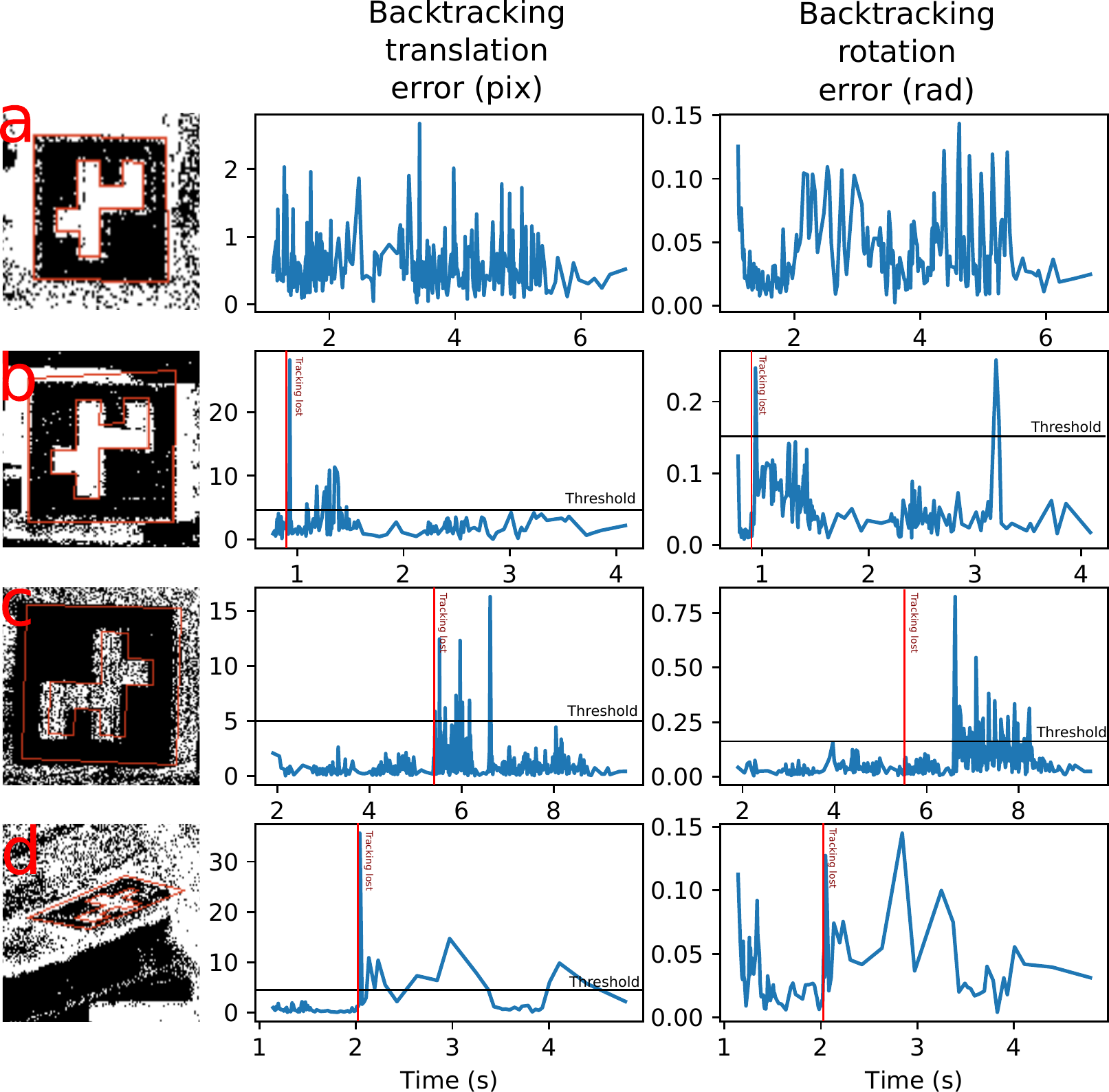}
\caption{Backtracking error signal for different common scenarios. For initial updates, the noisy signal is generated due to the refinement of the initial guess. We ignore the first N updates. In red is highlighted the detected lost tracking signal. (a) For a good initial estimation, the tracker follows easily the marker, the error signal remains in the reasonable range throughout the whole recording. (b,c) For a bad initial estimation, the tracker follows the object. However, if the tracker will not refine its initial pose, trackers are lost during the fast motion. (d) Partial occlusion or extreme rotation (small area of the marker) in the trajectory. In this case, if the tracker does not catch up on next updates, the tracker tends to stay in the rotated position.}
\label{backtrack}
\end{figure}

To show the limitations of our tracking algorithm, We present in Figure~\ref{backtrack} standard fail cases of the tracking, as well as verification signal for those trajectories. As shown, we can use the error signal to determine if the tracking was lost.

\subsection{Limits}
\begin{figure}[]
\centering
\includegraphics[scale=0.6]{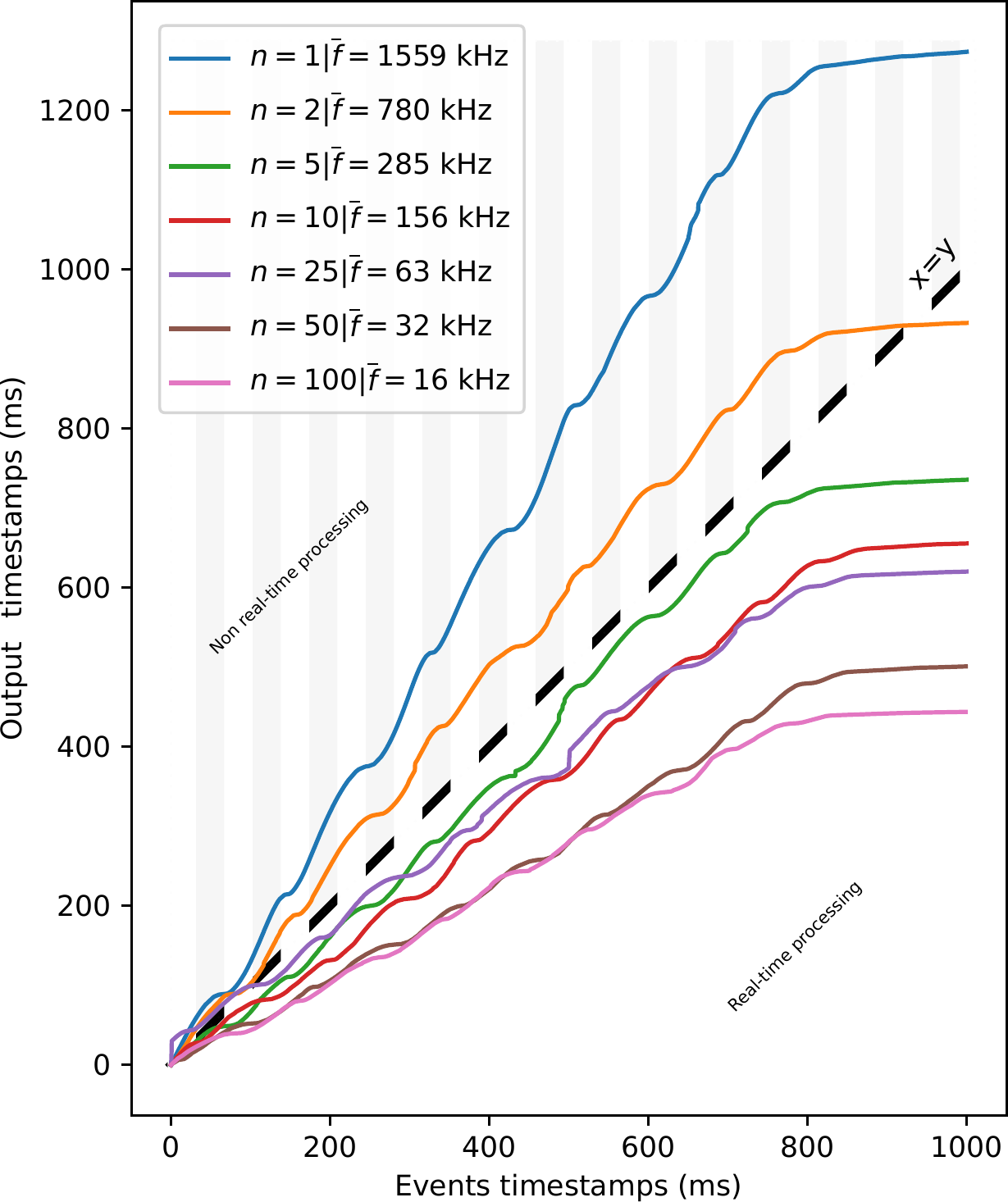}
\caption{Performance comparison for different update rates parameters. We here represent the event camera's timestamp against its estimated pose timestamp. Real-time domain lies below the x=y line.}
\label{limits}
\end{figure}

In order to visualize the trade-off between the signal rate and the performance, Figure~\ref{limits} presents the impact of different update rates on the computation time. Data for this figure has been collected processing the same recording with different update rates ($n$). Real-time performance is achieved for every point below the $x=y$ line; the pose output timestamp is then lower than the events timestamp. As we can see, for every $n<5$, a real-time operation cannot be achieved, due to heavy computational load. The maximum real-time update rate is here 156 kHz.
It has to be mentioned that a smaller update rate parameter will result in an increased delay, due to an heavier computational load. Also, a bigger update rate parameter will lower the tracking accuracy, while, however, improving robustness for non-perfect hyper-parameters. 
This analysis suggests that, given the shown margins, our approach will be viable for higher resolution sensors. Due to the parallel architecture of our solution, multiple marker tracking can be achieved using 2 threads per marker, without any delay increase. For our setup (8 cores processor), we were able to track up to 5 markers at the same time.

\par

\section{CONCLUSION AND FUTURE WORK}
\label{conclusion}

In this paper, we presented a novel approach for real-time fiducial marker tracking, using an event-based sensor. As proven by our experiments, the proposed method can be applied to tasks including high speed motion, as well as challenging lighting conditions in unconstrained environments. 
Our approach is able to achieve a real-time consistent tracking at an update rate up to 156~kHz, while maintaining an end-to-end latency lower than 3~ms.
Given a proper initial estimation, we demonstrated that we can achieve a long-term tracking. Nevertheless, further work should focus on improving the initial detection of the marker, as well as the tracker's robustness. More detailed work on the data flow management would also allow a further decrease of the overall latency.

We anticipate this work to become a building block for high-speed robotic applications requiring real-time visual feedback for closing the perception-action loop.
Exciting technological improvements of event-based sensors, especially in terms of spatial resolution, will bridge the accuracy gap that is still present nowadays.

\bibliographystyle{IEEEtran}
\bibliography{ref}

\begin{thebibliography}{10}
\providecommand{\url}[1]{#1}
\csname url@samestyle\endcsname
\providecommand{\newblock}{\relax}
\providecommand{\bibinfo}[2]{#2}
\providecommand{\BIBentrySTDinterwordspacing}{\spaceskip=0pt\relax}
\providecommand{\BIBentryALTinterwordstretchfactor}{4}
\providecommand{\BIBentryALTinterwordspacing}{\spaceskip=\fontdimen2\font plus
\BIBentryALTinterwordstretchfactor\fontdimen3\font minus
  \fontdimen4\font\relax}
\providecommand{\BIBforeignlanguage}[2]{{%
\expandafter\ifx\csname l@#1\endcsname\relax
\typeout{** WARNING: IEEEtran.bst: No hyphenation pattern has been}%
\typeout{** loaded for the language `#1'. Using the pattern for}%
\typeout{** the default language instead.}%
\else
\language=\csname l@#1\endcsname
\fi
#2}}
\providecommand{\BIBdecl}{\relax}
\BIBdecl

\bibitem{prop_new}
T.~Finateu, A.~Niwa, D.~Matolin, K.~Tsuchimoto, A.~Mascheroni, E.~Reynaud,
  P.~Mostafalu, F.~Brady, L.~Chotard, F.~LeGoff, H.~Takahashi, H.~Wakabayashi,
  Y.~Oike, and C.~Posch, ``5.10 a 1280×720 back-illuminated stacked temporal
  contrast event-based vision sensor with 4.86um pixels, 1.066geps readout,
  programmable event-rate controller and compressive data-formatting
  pipeline,'' in \emph{2020 IEEE International Solid- State Circuits Conference
  - (ISSCC)}, 2020, pp. 112--114.

\bibitem{sam_new}
Y.~Suh, S.~Choi, M.~Ito, J.~Kim, Y.~Lee, J.~Seo, H.~Jung, D.-H. Yeo,
  S.~Namgung, J.~Bong, S.~Yoo, S.-H. Shin, D.~Kwon, P.~Kang, S.~Kim, H.~Na,
  K.~Hwang, C.~Shin, J.-S. Kim, P.~K.~J. Park, J.~Kim, H.~Ryu, and Y.~Park, ``A
  1280×960 dynamic vision sensor with a 4.95-um pixel pitch and motion
  artifact minimization,'' in \emph{2020 IEEE International Symposium on
  Circuits and Systems (ISCAS)}, 2020, pp. 1--5.

\bibitem{cele}
S.~Chen and M.~Guo, ``Live demonstration: Celex-v: A 1m pixel multi-mode
  event-based sensor,'' in \emph{2019 IEEE/CVF Conference on Computer Vision
  and Pattern Recognition Workshops (CVPRW)}, 2019, pp. 1682--1683.

\bibitem{benchmarking_pose_estimation}
\BIBentryALTinterwordspacing
A.~Hietanen, J.~Latokartano, A.~Foi, R.~Pieters, V.~Kyrki, M.~Lanz, and J.-K.
  Kämäräinen, ``Benchmarking pose estimation for robot manipulation,''
  \emph{Robotics and Autonomous Systems}, vol. 143, p. 103810, 2021. [Online].
  Available:
  \url{https://www.sciencedirect.com/science/article/pii/S0921889021000956}
\BIBentrySTDinterwordspacing

\bibitem{bop}
T.~Hodan, M.~Sundermeyer, B.~Drost, Y.~Labbe, E.~Brachmann, F.~Michel,
  C.~Rother, and J.~Matas, ``Bop challenge 2020 on 6d object localization,'' in
  \emph{ECCV Workshops}, 2020.

\bibitem{apriltag}
E.~Olson, ``Apriltag: A robust and flexible visual fiducial system,'' in
  \emph{2011 IEEE International Conference on Robotics and Automation}.\hskip
  1em plus 0.5em minus 0.4em\relax IEEE, 2011, pp. 3400--3407.

\bibitem{stag}
B.~Benligiray, C.~Topal, and C.~Akinlar, ``Stag: A stable fiducial marker
  system,'' \emph{Image and Vision Computing}, vol.~89, pp. 158--169, 2019.

\bibitem{cctag}
L.~Calvet, P.~Gurdjos, C.~Griwodz, and S.~Gasparini, ``{Detection and Accurate
  Localization of Circular Fiducials under Highly Challenging Conditions},'' in
  \emph{{Proceedings of the 2016 IEEE Conference on Computer Vision and Pattern
  Recognition (CVPR)}}, Las Vegas, United States, Jun. 2016, pp. 562 -- 570.

\bibitem{aruco}
\BIBentryALTinterwordspacing
``Speeded up detection of squared fiducial markers,'' \emph{Image and Vision
  Computing}, vol.~76, pp. 38--47, 2018. [Online]. Available:
  \url{https://www.sciencedirect.com/science/article/pii/S0262885618300799}
\BIBentrySTDinterwordspacing

\bibitem{markerdetectioneb}
H.~Sarmadi, R.~Muñoz-Salinas, M.~A. Olivares-Mendez, and R.~Medina-Carnicer,
  ``Detection of binary square fiducial markers using an event camera,''
  \emph{IEEE Access}, vol.~9, pp. 27\,813--27\,826, 2021.

\bibitem{optitrack}
\url{OptiTrack - https://optitrack.com/}, accessed: 2021-09-13.

\bibitem{indoor}
\BIBentryALTinterwordspacing
A.~Morar, A.~Moldoveanu, I.~Mocanu, F.~Moldoveanu, I.~E. Radoi, V.~Asavei,
  A.~Gradinaru, and A.~Butean, ``\BIBforeignlanguage{eng}{A comprehensive
  survey of indoor localization methods based on computer vision},''
  \emph{\BIBforeignlanguage{eng}{Sensors (Basel, Switzerland)}}, vol.~20,
  no.~9, p. 2641, May 2020. [Online]. Available:
  \url{https://pubmed.ncbi.nlm.nih.gov/32384605}
\BIBentrySTDinterwordspacing

\bibitem{artoolkit}
D.~Wagner and D.~Schmalstieg, ``Artoolkitplus for pose tracking on mobile
  devices,'' 01 2007.

\bibitem{swarms}
L.~Mateos, ``Apriltags 3d: Dynamic fiducial markers for robust pose estimation
  in highly reflective environments and indirect communication in swarm
  robotics,'' 01 2020.

\bibitem{artag}
M.~Fiala, ``Artag, a fiducial marker system using digital techniques,'' in
  \emph{2005 IEEE Computer Society Conference on Computer Vision and Pattern
  Recognition (CVPR'05)}, vol.~2, 2005, pp. 590--596 vol. 2.

\bibitem{robustaruco}
V.~M. Mondejar-Guerra, S.~Garrido-Jurado, R.~Munoz-Salinas, M.~Marín-Jimenez,
  and R.~Medina-Carnicer, ``Robust identification of fiducial markers in
  challenging conditions,'' \emph{Expert Systems with Applications}, vol.~93,
  10 2017.

\bibitem{robustaruco1}
B.~Li, J.~Wu, X.~Tan, and B.~Wang, ``Aruco marker detection under occlusion
  using convolutional neural network,'' in \emph{2020 5th International
  Conference on Automation, Control and Robotics Engineering (CACRE)}, 2020,
  pp. 706--711.

\bibitem{blurresistant}
M.~G. Prasad, S.~Chandran, and M.~S. Brown, ``A motion blur resilient fiducial
  for quadcopter imaging,'' in \emph{2015 IEEE Winter Conference on
  Applications of Computer Vision}, 2015, pp. 254--261.

\bibitem{runetag}
F.~Bergamasco, A.~Albarelli, L.~Cosmo, E.~Rodolà, and A.~Torsello, ``An
  accurate and robust artificial marker based on cyclic codes,'' \emph{IEEE
  Transactions on Pattern Analysis and Machine Intelligence}, vol.~38, no.~12,
  pp. 2359--2373, 2016.

\bibitem{pitag}
F.~Bergamasco, A.~Albarelli, and A.~Torsello, ``Pi-tag: A fast image-space
  marker design based on projective invariants,'' \emph{Machine Vision and
  Applications}, vol.~24, 08 2013.

\bibitem{topotag}
\BIBentryALTinterwordspacing
G.~Yu, Y.~Hu, and J.~Dai, ``Topotag: {A} robust and scalable topological
  fiducial marker system,'' \emph{{IEEE} Trans. Vis. Comput. Graph.}, vol.~27,
  no.~9, pp. 3769--3780, 2021. [Online]. Available:
  \url{https://doi.org/10.1109/TVCG.2020.2988466}
\BIBentrySTDinterwordspacing

\bibitem{detection1mpix}
\BIBentryALTinterwordspacing
E.~Perot, P.~de~Tournemire, D.~Nitti, J.~Masci, and A.~Sironi, ``Learning to
  detect objects with a 1 megapixel event camera,'' \emph{CoRR}, vol.
  abs/2009.13436, 2020. [Online]. Available:
  \url{https://arxiv.org/abs/2009.13436}
\BIBentrySTDinterwordspacing

\bibitem{rgbde}
E.~Dubeau, M.~Garon, B.~Debaque, R.~d. Charette, and J.-F. Lalonde, ``Rgb-d-e:
  Event camera calibration for fast 6-dof object tracking,'' in \emph{2020 IEEE
  International Symposium on Mixed and Augmented Reality (ISMAR)}, 2020, pp.
  127--135.

\bibitem{reconstruction}
\BIBentryALTinterwordspacing
F.~Paredes{-}Vall{\'{e}}s and G.~C. H.~E. de~Croon, ``Back to event basics:
  Self-supervised learning of image reconstruction for event cameras via
  photometric constancy,'' \emph{CoRR}, vol. abs/2009.08283, 2020. [Online].
  Available: \url{https://arxiv.org/abs/2009.08283}
\BIBentrySTDinterwordspacing

\bibitem{filtering}
A.~Khodamoradi and R.~Kastner, ``$\mathcal{O}(n)$-space spatiotemporal filter
  for reducing noise in neuromorphic vision sensors,'' \emph{IEEE Transactions
  on Emerging Topics in Computing}, vol.~9, no.~1, pp. 15--23, 2021.

\bibitem{particle_filter}
D.~Weikersdorfer and J.~Conradt, ``Event-based particle filtering for robot
  self-localization,'' in \emph{2012 IEEE International Conference on Robotics
  and Biomimetics (ROBIO)}, 2012, pp. 866--870.

\bibitem{slam}
H.~Kim, A.~Handa, R.~Benosman, S.-H. Ieng, and A.~Davison, ``Simultaneous
  mosaicing and tracking with an event camera,'' in \emph{Proceedings of the
  British Machine Vision Conference}.\hskip 1em plus 0.5em minus 0.4em\relax
  BMVA Press, 2014.

\bibitem{ebpose}
\BIBentryALTinterwordspacing
D.~Reverter~Valeiras, G.~Orchard, S.-H. Ieng, and R.~B. Benosman,
  ``Neuromorphic event-based 3d pose estimation,'' \emph{Frontiers in
  Neuroscience}, vol.~9, p. 522, 2016. [Online]. Available:
  \url{https://www.frontiersin.org/article/10.3389/fnins.2015.00522}
\BIBentrySTDinterwordspacing

\bibitem{ebpnp}
\BIBentryALTinterwordspacing
D.~Reverter~Valeiras, S.~Kime, S.-H. Ieng, and R.~B. Benosman, ``An event-based
  solution to the perspective-n-point problem,'' \emph{Frontiers in
  Neuroscience}, vol.~10, p. 208, 2016. [Online]. Available:
  \url{https://www.frontiersin.org/article/10.3389/fnins.2016.00208}
\BIBentrySTDinterwordspacing

\bibitem{lu_proofs}
C.-P. Lu, G.~Hager, and E.~Mjolsness, ``Fast and globally convergent pose
  estimation from video images,'' \emph{IEEE Transactions on Pattern Analysis
  and Machine Intelligence}, vol.~22, no.~6, pp. 610--622, 2000.

\bibitem{openeb}
\url{OpenEB, https://github.com/prophesee-ai/openeb}, accessed: 2021-09-13.

\bibitem{evk}
\url{Prophesee Sensor - Gen3 Evaluation Kit (640x480),
  https://https://www.prophesee.ai/event-based-evk-3/}, accessed: 2021-09-13.

\bibitem{survey}
G.~Gallego, T.~Delbruck, G.~M. Orchard, C.~Bartolozzi, B.~Taba, A.~Censi,
  S.~Leutenegger, A.~Davison, J.~Conradt, K.~Daniilidis, and D.~Scaramuzza,
  ``Event-based vision: A survey,'' \emph{IEEE Transactions on Pattern Analysis
  and Machine Intelligence}, pp. 1--1, 2020.

\end{thebibliography}

\end{document}